# Weakly Contrastive Learning via Batch Instance Discrimination and Feature Clustering for Small Sample SAR ATR

Yikui Zhai [1], *Member, IEEE,* Wenlve Zhou [1], Bing Sun [2,*], Jingwen Li [2], Qirui Ke [1], Zilu Ying [1], Junying Gan [1], Chaoyun Mai [1], Ruggero Donida Labati [3], *Member, IEEE,* Vincenzo Piuri [3], *Fellow, IEEE* and Fabio Scotti [3], *Senior Member, IEEE*

*Abstract*—In recent years, impressive performance of deep learning technology has been recognized in Synthetic Aperture Radar (SAR) Automatic Target Recognition (ATR). Since a large amount of annotated data is required in this technique, it poses a trenchant challenge to the issue of obtaining a high recognition rate through less labeled data. To overcome this problem, inspired by the contrastive learning, we proposed a novel framework named *Batch Instance Discrimination and Feature Clustering (BIDFC)*. In this framework, different from that of the objective of general contrastive learning methods, embedding distance between samples should be moderate because of the high similarity between samples in the SAR images. Consequently, our flexible framework is equipped with adjustable distance between embedding, which we term as weakly contrastive learning. Technically, instance labels are assigned to the unlabeled data in per batch and random augmentation and training are performed *few* times on these augmented data. Meanwhile, a novel *Dynamic-Weighted Variance loss (DWV loss)* function is also posed to cluster the embedding of enhanced versions for each sample. Experimental results on the moving and stationary target acquisition and recognition (MSTAR) database indicate a 91.25% classification accuracy of our method fine-tuned on only 3.13% training data. Even though a linear evaluation is performed on the same training data, the accuracy can still reach 90.13%. We also verified the effectiveness of BIDFC in OpenSarShip database, indicating that our method can be generalized to other datasets. Our code is avaliable at: https://github.com/Wenlve-Zhou/BIDFC-master.

*Index Terms*—Deep learning, weakly contrastive learning, batch instance discrimination, dynamic-weighted variance loss, small sample SAR ATR.

## I. INTRODUCTION

Automatic target recognition (ATR) tends to classify real-world targets into several categories with images from imaging sensor. Currently, synthetic aperture radar (SAR) ATR, has aroused tremendous interest from numerous researchers, which can mainly be attributed to its effectiveness in reducing sensitivity to weather conditions and long standoff as SAR system can generate an image of the reflectivity distribution of the surface being observed [1]-[5]. Nevertheless, the limitations of SAR ATR still consecutively hinder its development, such as serious speckle noise, severe geometric distortion, critical structural defects, and low angle sensitivity.

Fortunately, deep learning is capable of automatically learning highly hierarchical image features from large-scale datasets; therefore, the techniques have been widely applied in remote sensing data analysis in recent years [6], [7], effectually alleviating the defects of SAR ATR. For example, Huang *et al.* [6] designed an assembled Convolutional Neural Network (CNN) equipped with a classification branch and a reconstruction branch, as well as a feedback bypass additionally, which proved 99.05% accuracy on the MSTAR dataset. In addition, a new all-convolutional architecture (A-ConvNets) only consisting of sparsely connected layers, but without fully connected layers being used, was proposed by Chen *et al.* [7], which engendered an average accuracy of 99% on the MSTAR dataset. Thanks to the neural network technology, researchers have obtained impressive results on the SAR ATR. Nonetheless, two major problems still exist in this area: (i) Despite a much smaller scale of SAR image datasets compared with that of the other ones, experiments conducted on SAR image datasets can still sustain considerable performance with the help of neural

1. Y. Zhai, W. Zhou, Q. Ke, Z. Ying, J. Gan, and C. Mai are with the Department of Intelligent Manufacturing, Wuyi University, Jiangmen 529020, China (e-mail: yikuizhai@163.com; wenlvezhou@163.com; 17683921638@163.com; ziluy@163.com; junyinggan@163.com; maichaoyun@foxmail.com).

2. B. Sun and J. Li are with the School of Electronics and Information Engineering, Beihang University, Beijing 100191, China (e-mail: bingsun@buaa.edu.cn; lijingwen@buaa.edu.cn).

3. R. D. Labati, V. Piuri, and F. Scotti are with Departimento di Information, Universita, Degli Studi di Milano, via Celoria 18, 20133 Milano (MI), Italy (e-mail: Ruggero.Donida@unimi.it; vincenzo.piuri@unimi.it; fabio.scotti@unimi.it )

* Correspondence: bingsun@buaa.edu.cn; Tel.: +010-8233-8670

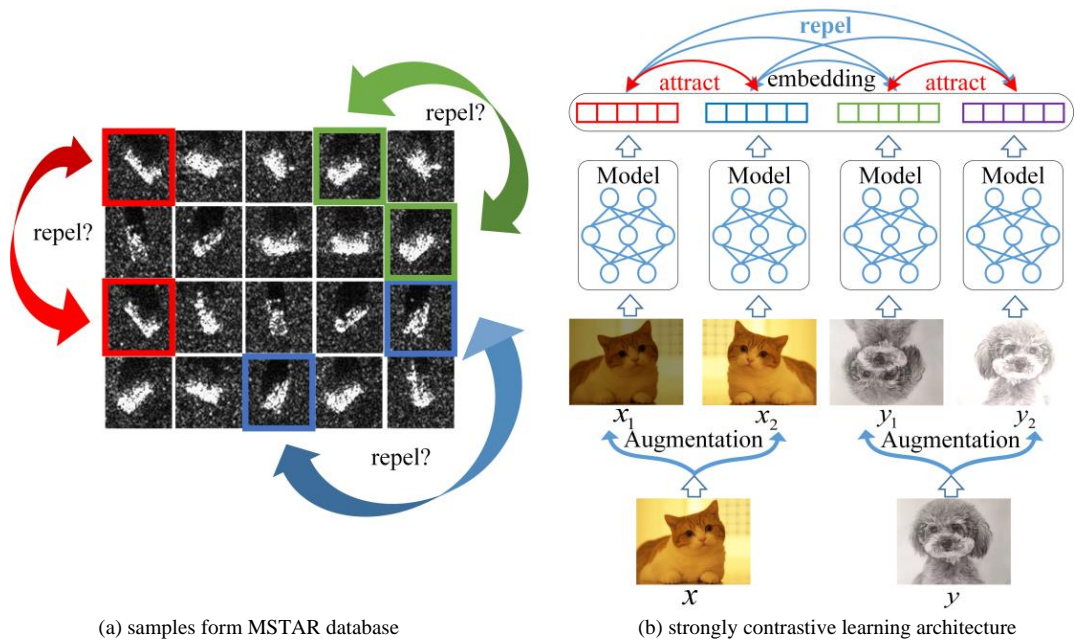

(a) samples form MSTAR database　　　　　　　　　　(b) strongly contrastive learning architecture

Fig. 1. Visualization of MSTAR dataset and strongly contrastive learning architecture. (a): There are 10 categories in the MSTAR dataset. We randomly select two pictures from each category in this dataset, where three similar image pairs are annotated with boxes of the same color. (b): Random augmentation is performed on per image twice, with the feature extracted from the general ConvNet. The embedding distance of the same image should be attracted, while those of the different ones should be repelled.

network technology. This proves that smaller amount of training data is possible if more tricks are applied. (ii) The recent state-of-the-art (SOTA) methods of SAR ATR will limit the deployment of the ATR system for they aim at the combination of network design and transfer learning [6]-[9]. Consideration should not only be given to matching model with suitable size but also to reducing the discrepancy of feature representation between the source domains and target domains when utilizing transfer learning, since its strategy will vary with the network architecture and the source domain.

In the field of computer vision, various researchers are committed to exploring general methods, rather than only focusing on network architecture, thus drawing forth copious outstanding methods. Among them, the most effective one by now belongs to self-supervised learning (SSL) that mainly uses pretext tasks to dig out its supervised information from large-scale non-label databases, and the network is trained to learn valuable representations for downstream tasks [10]-[13]. The operation of SSL includes unsupervised feature learning for unlabeled samples and fine-tuned/linear evaluation based on small amounts of annotated data. In self-supervised learning, these methods based on classification tasks can be divided into two categories: context learning and contrastive learning. Context learning mainly exploits the structural information of the data, artificially constructs tags based on the structural information and then trains it like supervised learning [14]-[18]. However, when a low degree of correlation appears in pretext tasks, the performance of context learning to learn a good visual representation for the downstream tasks will drop accordingly. Discriminative approaches based on contrastive learning in the latent space have recently revealed great performance [19]-[23], achieving SOTA in the computer vision tasks, among which the most outstanding representatives are Chen *et al.* [16] who proposed a framework Simclr fine-tuned on 1% of the ImageNet database [24] outperforming AlexNet with 100× fewer labels. In the mainstream framework (Fig.1b), a complex augmentation is deployed on the data to obtain correlated views $x_1$, $x_2$, and on a neural network to encode them respectively to obtain the embedding. Intuitively, $x_1$ and $x_2$ as the derivation of $x$, their embedding should be similar, which is called "attract". Conversely the embedding from different samples e.g. $x_1$ and $y_1$, they share extremely low similarity, which is called "repel". These two processes are regarded as supervision information to optimize the network so that the model can learn useful semantic representation.

There is no denying that successful results of the architecture have been testified on optical image datasets with diverse data samples, but the story goes differently on SAR ATR. Since Intra-class samples of the SAR dataset possess high similarity, such as the MSTAR dataset (In Fig.1a), it is not required to have a huge discrepancy in embedding between different samples for representation learning of SAR images in SAR ATR. Howbeit the mainstream contrastive learning framework, which we define as strongly contrastive learning (SCL) in this paper, commit to obtain significant differences on embedding between different samples. Hence, to counter the issue of invalid application of the general contrastive learning framework to SAR ATR, we pioneer with a novel contrastive learning framework called *Batch Instance Discrimination with Feature Clustering (BIDFC)*, and we define this framework as weakly contrastive learning (WCL) **because it no longer requires a huge discrepancy in the similarity of the embedding space of different samples, perfectly explaining the "weakly" in its name WCL.**

Our main contribution can be summarized as follows:

- A novel contrastive learning framework BIDFC for unsupervised feature learning is presented. It can effectively overcome the small sample effect and alleviate the issue of noise and fuzzy disturbance in SAR images.
- The batch instance discrimination as the "repel" part is proposed to adjust the intensity of contrastive learning, enabling our method to switch flexibly between SCL and WCL.
- A novel *Dynamic-Weighted Variance loss (DWV loss)* function for feature clustering is projected as the "attract" part in our framework, realizing contrastive learning of multiple samples.
- The SOTA methods are enforced on the MSTAR dataset including transfer learning, semi-supervised learning, self-supervised learning, and unsupervised representation learning, compared with which BIDFC pertains to state of-the-art on MSTAR benchmark. Besides, the generalization of BIDFC is verified on OpenSARShip dataset.

The rest of the paper is organized as follows: After a brief introduction to SAR ATR using deep learning and self-supervised learning in Section II, the proposed architecture is elaborated in Section III. In Section IV, experimental results are displayed to discuss and verify the effectiveness of our method, followed by the complete conclusion in Section V.

## II. RELATED WORK

In this paper, we intend to delve into self-supervised learning on small samples SAR target recognition, and yet no researchers have ever used SSL to study SAR ATR while few methods that focus on training the neural network on small samples SAR dataset can be found. Consequently, we will introduce some classical methods on SAR ATR using deep learning techniques and several related researches about context learning and contrastive learning.

### A. SAR ATR In Deep Learning

Morgan [25] as the forerunners of using ConvNet for SAR target recognition, their framework held an average accuracy of 92.25% even without mature neural networks [26]-[32] and training skills [33], [34] that speed up convergence and avoid overfitting. This reflects the little influence of the limited data on the framework's performance, which implies the plausibility to train a network with tricks and special structure for small samples in SAR ATR. Gao *et al.*[35] combined CNN with support vector machine (SVM), which trained with an improved loss function. They achieved 99% average accuracy on the MSTAR dataset. Yue *et al.* [36] proposed a novel semi-supervised learning method. In the supervised learning compoment, they used Cross-Entropy for training, and in the unsupervised part, they proposed a new e linear discriminant analysis method for unlabeled sample loss calculation. Zhao *et.al.* [37] suggested a novel CNN architecture based on the highway unit [38], in which 30% random sampled data from the MSTAR dataset was extracted to form the collection of a small sample. Despite of its impressive accuracy at 94.97%, the amount of extracted sampled data tends to be unbearably excessive for small sample learning. This leads to the unsatisfying situation that the recognition rate will decline sharply, due to the lack of different rotation angles with less than 30% training data. In section IV, our experiments certify the mild relationship between the diversity of angles in SAR images and the maintainance of considerable accuracy when trained on the few-shot database (In Table X). The main reason for angle insensitivity can be attributed to the absence of a good visual representation. In recent years, the most successful solution to this problem is self-supervised learning.

### B. Context Learning

In computer vision area, Doersch *et al.* [14] primarily drew researchers' attention to SSL by extracting two random but non-overlapping patches pair from the same image and training a convolutional neural network to predict the relative location of the two patches. After that, Noroozi *et al.* [15] made further improvements through designing a jigsaw puzzle as the pretext, which could divide an image into 9 patches, then followed by training a neural network to predict the correct order. Afterwards, Gidaris *et al.* [16] utilized CNN to predict the angle of images that have been performed $0°, 90°, 180°, 270°$ operations. Besides, in light of the unfavorable flexibility of self-supervised learning, Beyer *et al.* [17] combined semi-supervised learning with self-supervised learning to construct a new framework, which was mainly composed of two branches. One performed supervised learning that classified the objects into several categories while the other was in charge of self-supervised learning, predicting the angle of each image. Moreover, Zhang *et al.* [18] is also worth mentioning for they regarded colorful image colorization as pretext tasks. They first gray-scaled the colorful images and then utilized fully convolutional neural network to colorize the grayscale images. Unfortunately, owing to various rotation angles, serious speckle noise, detrimental geometric distortion, and unfavorable grayscale characteristics of SAR images, the majority of context learning methods fail to suit unsupervised feature learning based on SAR images. Therefore, we shift our sight to another branch of SSL: contrastive learning.

### C. Contrastive Learning

Contrastive learning can be concluded as two aspects: reducing the distance of embedding between similar samples and increasing the distance of embedding between different samples. Researchers refer to the former one as positive sample, and the latter as negative samples. In the study of Hadsell *et al.* [19], a set of data similar to $x$ for each input sample $x$ was produced to facilitate the application of CNN so as to create its embedding. Meanwhile, a distance-based contrastive loss function was deployed for the optimization of this model, by means of reducing the euclidean distance between positive samples and increasing the distance between negative samples. As for Wu *et al.* [20], they took each sample in the dataset as a category and used a "memory bank" to store its feature embedded from the model. In the next epoch, they carried out contrastive learning on the newly encoded features and the embedding stored in the memory bank. As a result of the positive correlation between the batch size and the performance of contrastive learning, Chen *et al*. [21] explored a new means to improve the performance of contrastive learning from the perspective of engineering. They designed an augmentation combination for ImageNet database that was conducive to unsupervised representation learning, as well as utilized the framework of Fig.1b to optimize the network. It is worth noting that the appliance of 128 tensor processing units (TPUs) is complemented with the need of larger batch size (e.g. 2048, 4096) for training. To change the luxury of computation, He *et al*. [22] followed the memory bank technique, applied queue to store the feature encoded by CNN and also employed the momentum encoder to solve the huge discrepancy between features in the memory bank and newly encoded features----a fresh method named Moco. In [23], Moco realized advancement with the efforts from Chen *et al*. who applied the augmentation and project head from [21], successfully outperforming the Simclr.

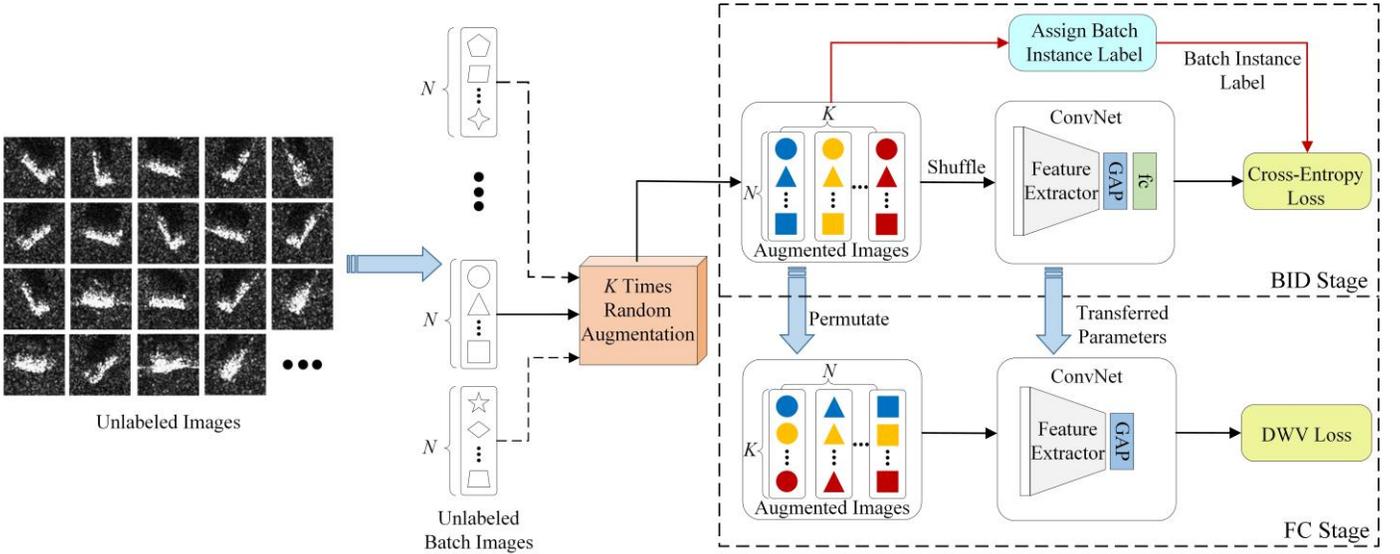

Fig. 2. The pipeline of weakly contrastive learning approach. The different shapes in the figure represent different instance samples, and the same color indicate these samples transformed from the same augmentation. In the BID stage, batch instance labels are assigned to $K$ batches of enhanced samples firstly. Then features encoded by ConvNet and batch instance labels are trained with cross-entropy loss. In the FC stage, $K$ batches of enhanced samples are permutate and reencode by the feature extractor with GAP which obtained in the BID stage. These reencoded features are trained with DWV Loss.

On account of numerous similar samples in the SAR imagedatabase, the strongly contrastive learning framework mentioned above has already demonstrated its unsuitability for unsupervised feature learning in the field. For this reason, we are committed to exploring the specialized framework and augmentation combination to contrastive learning for the SAR database in this article.

## III. METHODS

Our algorithm is mainly composed of two parts: batch instance discrimination (BID) and feature clustering (FC), which correspond to "repel" and "attract" in contrastive learning respectively. The principle of common contrastive learning framework to utilize correlated views from random augmentation has been adopted for our framework, which will be evaluated through fine-tuning and linear evaluation. For this reason, augmentations for unsupervised feature learning based on SAR images and evaluation methods will be investigated here also. The framework of this algorithm is demonstrated in Fig.2.

### A. Batch Instance Discrimination

Batch instance discrimination is inspired by instance discrimination brought up by Wu *et al*. [20]. We therefore formulated instance discrimination before introducing our method. Instance discrimination treats each sample in the entire database as one category (e.g. ImageNet has 14 million samples, so there are 14 million corresponding categories). Aiming to mitigate the influence of the tremendous amount of parameter and costly computation resulting from fully connected layer, a non-parametric classifier was proposed by them, written as:

$$P(i \mid v) = \frac{\exp(v_i^T v)}{\sum_{j=1}^{n} \exp(v_j^T v)} \tag{1}$$

Where $v$ refers to the embedding from the fully connected layer and $V = \{v_j\}$ represents the memory bank. Due to the fact that instance discrimination replaces the embedding of the memory bank to the weight learned from the neural network, the essence of this approach lies in strongly contrastive learning. During the optimization of the objective function, the embedding would become closer on a similar sample (i.e. the augmented versions of the image), while further on different samples with $P(i \mid v)$ close to 1. This situation is not beneficial to the unsupervised feature learning on the SAR database.

In response to this problem, we propose the batch instance-level discrimination to reduce expensive computation cost. The core of BID is similar to supervised learning in that it assigns an instance-level label to each sample in per batch for the model training. During the process of batch instance-level discrimination, two points deserve our attention. First, the batch instance label of each sample must be different in each epoch.

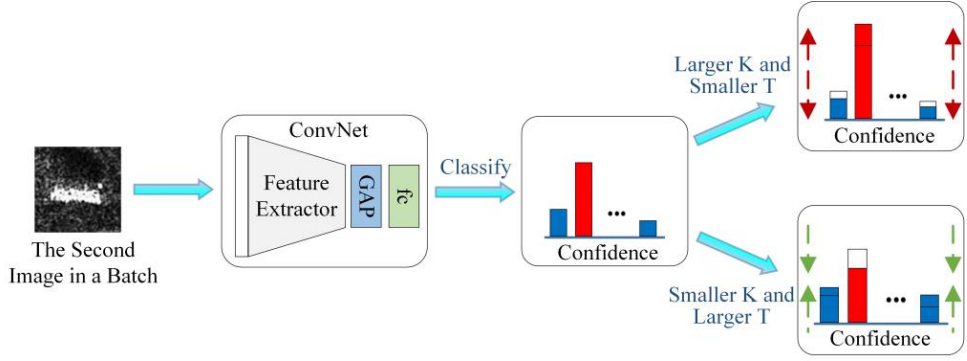

Fig. 3. Illustration of the influence of augmentation times $K$ and temperature coefficient $T$ on confidence.

Otherwise, it will just train a ConvNet to randomly divide all data into several categories that easily converge, but unable to learn the useful semantic representation. Second, since differences must exist in the label of the samples in each epoch, it is incapable of making the model converging in a normal training way. So receiving the inspiration from Hermann Ebbinghaus, an experimental psychologist, a new method is employed to model training. For a batch of samples, we perform $K$ times augmentation on each sample in the batch so as to obtain $K$ batches of augmented data, with the need to be trained. This operation administers to the model to "remember" the features of these pictures via enabling the model "watch" the batch of samples $K$ times. Besides, after the data of each batch has been trained for $K$ times, we will randomly initialize the weight of the fully connected layer i.e. softmax classifier, which can help the model converge better.

Providing a batch of images $x_1, ..., x_n$ in $N$ classes and their embedding $v_1, ..., v_n$ from the fully connected layer, we replace traditional softmax classifier to non-parametric classifier Eq. (1), and the probability of $P(i \mid v)$ becomes:

$$P(i \mid v) = \frac{\exp(w_i^T v)}{\sum_{j=1}^{n} \exp(w_j^T v)} \tag{2}$$

where $w_i$ pertains to be a weight for class $i$ from the fully connected layer. Using the weights of the model for batch instance discrimination can adaptively adjust the distance between embedding assisted by model training, and indirectly repel the features between samples. Meanwhile, we add a temperature coefficient $T$ [39] to Eq. (2) for better control the strength of contrastive learning, as

$$P(i \mid v) = \frac{\exp(w_i^T v / T)}{\sum_{j=1}^{n} \exp(w_j^T v / T)} \tag{3}$$

The temperature coefficient is correlative to confidence. When $T$ is larger, the probability distribution is smoother, which can lead to smaller discrepancy between embedding in the feature space and vice versa. $K$ can also be adopted to control intensity of the model. The more images $K$ "watches", the higher confidence of the model and the larger feature distance can be obtained. Consequently, two parameters $K$ and $T$ can enable flexible changes between SCL and WCL in BIDFC. The influence of hyperparameters $K$ and $T$ is shown in Fig.3.

Although the strength of contrastive learning can be adjusted with the control of hyperparameters $K$ and $T$, inappropriate hyperparameters (e.g. too small $K$ and too large $T$) can still weaken the intensity of contrastive learning, resulting in damages of the visual representation learning. In the ablation study (section IV), we will find out the optimized augmentation times and temperature coefficient for BIDFC through experiments. At last, the objective is to minimize the negative log-likelihood over the training set, and we call it Cross-Entropy loss in this article, written as

$$L_{ce} = -\frac{1}{n} \sum_{i=1}^{n} \log \frac{\exp(w_i^T v / T)}{\sum_{j=1}^{n} \exp(w_j^T v / T)} \tag{4}$$

### B. Feature Clustering with Dynamic-Weighted Variance Loss

Batch instance discrimination succeeded in settling two problems, one of which is the huge computation required by the pretext task instance discrimination; the other one of which refers to the unsatisfying intensity of contrastive learning. Even though the problems mentioned ahead have been countered, another new problem arises when applying $K$ times augmentation to representation learning. General contrastive learning framework [21], [22] only performs two times augmentation on each image and calculates the similarity between two samples. But for our proposed method BIDFC, calculation times should be $K$, far more than the two in general contrastive learning framework.

A simple and straightforward solution is to calculate and minimize the similarity between every two samples. But this approach appears to be pricey, especially when the $K$ is large. After applying $K$ times augmentation on the samples in a batch, we can get an augmented dataset: $I = \{I_{11}, I_{12}, ..., I_{nk}\}$. To solve the problem in a relatively small calculation, we adopted a novel loss function named DWV loss, written as:

$$L_{dwv} = \text{weight}(t) \frac{1}{nk} \sum_{i=1}^{n} \sum_{j=1}^{k} (z_i^k - \mu)^2 \tag{5}$$

$$\mu = \frac{1}{k} \sum_{j=1}^{k} z_i^k \tag{6}$$

where $z_i^k$ stands for an embedding from the global average pooling (GAP) layer, the training epochs denoted as $t$, and weight($t$) represents the hyperparameter to adjust the DWV loss. The illustration of feature clustering is displayed in Fig.4.

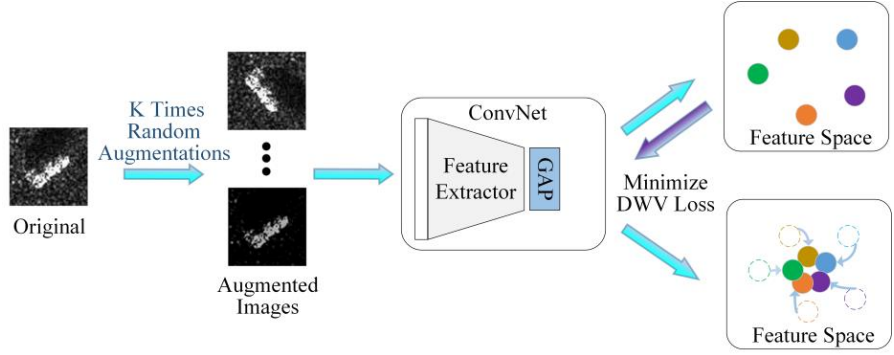

Fig. 4.  Illustration of feature clustering. The balls of different colors represent the embedding of the enhanced samples encoded from the GAP in the feature space.

The proper scheduling of the weight($t$) in the Eq. (5) should be attached critical importance to because it closely links with the network performance. When the model cannot distinguish between different samples, the ConvNet will quickly make it close to 0 if we use a large weight to minimize the DWV loss. This shortcut is to make the model encode any sample close to 0, and it will be seriously harmful to unsupervised feature learning. On the contrary, we also cannot benefit from unlabeled data if the weight appears to be too small. Inspired by [40], [41], we define the weight as a slowly increasing function of time $t$ . Moreover, we hope that the value of weight will remain at zero from $0$ to $T_1$ so that the network can distinguish different samples via batch instance discrimination. While the training is stable, the value will remain unchanged i.e. remain the maximum value of weight($t$) , and this stable moment is defined as $T_2$ . In this part, we apply a linear-schedule function as weight (Eq. (7)), read as:

$$linear - schedule = \begin{cases} 0 & t < T_1 \\ \dfrac{t - T_1}{T_2 - T_1} \alpha_f & T_1 \le t < T_2 \\ \alpha_f & T_2 \le t \end{cases} \quad (7)$$

The above are the two major components of BIDFC. During training process, assisted by exponential moving average (EMA) to update the parameters, a more stable model during the training process and greater visual representation can be obtained. Defining the parameters of feature extractor as $\theta_t$ in the current epoch, $m$ as a momentum coefficient range from 0 to 1, and $\theta_{t-1}$ as the historical parameters of the model, we can update $\theta_t$ by:

$$\theta_t \leftarrow m\theta_{t-1} + (1-m)\theta_t \quad (8)$$

Algorithm1 provides the pseudo-code of BIDFC for this pretext task. For batch instance discrimination and feature clustering, their optimization is not performed simultaneously. We consider that when the network still cannot distinguish between different samples, feature clustering of the augmented version will make the model fall into the trivial solution and fail to learn semantic features.

---

**Algorithm 1** Pseudocode of BIDFC.

**Input:** epochs, batch size $N$, numbers of augmentation $K$, feature extractor with GAP $f_\theta$, fully connected layer $f_{fc}$, momentum coefficient $m$, random augmentation $A$, temperature coefficient $T$, weight($t$) for *DWV loss*.

**for all** $t \in \{1,...,epochs\}$ **do**
　**for** each minibatch **do**
　　Initialize the parameters of $f_{fc}$
　　**for all** $k \in \{1,...,K\}$ **do**
　　　draw K augmentation functions $a^k \sim A$
　　　**for** sampled minibatch $\{x_i\}_{i=1}^{N}$ **do**
　　　　**for all** $i \in \{1,...,N\}$ **do**
　　　　　assign instance label $y_i^k$ to $\tilde{x}_i^k$ where $y_i^k = i$
　　　　　# the k-th augmentation
　　　　　$\tilde{x}_i^k = a^k(x_i)$
　　　　　$h_i^k = f_{fc}^t\left(f_\theta^t\left(\tilde{x}_i^k\right)\right)$ # features from the fully connected layer
　　　　**end for**
　　　**end for**
　　　**define** $L_{ve} = -\dfrac{1}{KN}\sum_{k=1}^{K}\sum_{i=1}^{N} y_i^k \log \dfrac{\exp\left(h_i^k / \mathrm{T}\right)}{\sum_{j=1}^{n} \exp\left(h_j^k / \mathrm{T}\right)}$
　　　update networks $f_\theta^t$ and $f_{fc}^t$ to minimize $L_{ve}$
　　**end for**
　　**for all** $k \in \{1,...,K\}$ and $i \in \{1,...,N\}$ **do**
　　　$z_i^k = f_\theta^t\left(\tilde{x}_i^k\right)$ # features from global average pooling layer
　　**end for**
　　**define** $L_{dwv} = \dfrac{\text{weight}(t)}{KN}\sum_{i=1}^{N}\sum_{k=1}^{K}\left(z_i^k - \dfrac{1}{K}\sum_{k=1}^{K} z_i^k\right)^2$
　　update networks $f_\theta^t$ to minimize $L_{dwv}$
　**end for**
　# momentum update network
　$f_\theta^t = mf_\theta^{t-1} + (1-m)f_\theta^t$
**end for**

---

### C. Data Augmentation for BIDFC

Data augmentation has been widely used in supervised learning and unsupervised representation learning. In particular, Chen *et al.* [21] verified that visual representation could benefit from different augmentation combinations. However, it is a pity that this augmentation is unsuitable for feature learning of SAR images because it is based on the study of optical images. In this section, several augmentation methods will be introduced and discussed so as to locate the optimal augmentation combinations for unsupervised feature learning.

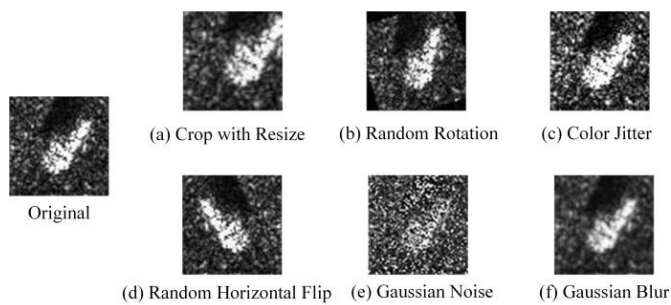

Fig. 5. Illustration of the studied data augmentation method. The original image is a sample from the MSTAR dataset, and (a)~(f) is the augmented version. Each operation can transform data randomly with some internal parameters (e.g. rotation degree).

When taking the variety of the scales of the target in SAR images into consideration, we introduce crop with resize (CR) to increase the adaptability of the model to the objects with different scales. Meanwhile, during the process of SAR images generation, the generation system is prone to bring in speckle noise and fuzzy disturbance. Hence, we employ Gaussian noise (GN) and Gaussian blur (GB) into unsupervised feature learning, hoping that the model can learn the invariance of noise and blur. To strengthen the richness of the augmentation combinations and the complexity of the pretext, three common data transformations are ushered in as well, named random rotation (RR), color jitter (CJ), and random horizontal flip (RHF) accordingly, whose operating outcomes are exhibited in Fig.5. In ablation study, the internal parameters of operations will be discussed and the exploration of the impact of each transformation will be conducted with the aim to learn the ideal augmentation combinations for representation learning. The experiments are shown in Table III.

### D. Fine-Tuned and Linear Evaluation

Fine-tuned and linear evaluation, as the measurement of the methods in this work, their concepts are raised first and foremost. Fine-tuned is a technology of transfer learning, which is extensively applied in academia and industry. In supervised learning, the model obtained by training on a database with a large amount of data (source domain) is called the pre-trained model, which serves to train in the target domain. Similarly, unsupervised feature learning performs feature learning operation on unlabeled samples and fine-tuned the model with a few labeled data as supervised learning does. In the fine-tuned phase, the weights of all layers will be trained, and equally the feature distribution on the feature space will be altered to obtain better performance.

Linear evaluation refers to a recent method used to evaluate unsupervised feature learning [20]-[22]. This method specializes in freezing the weight of the feature extractor and training only the fully connected layer [21], [22], or applying machine learning classifier [20] (e.g. KNN [42], Logistic

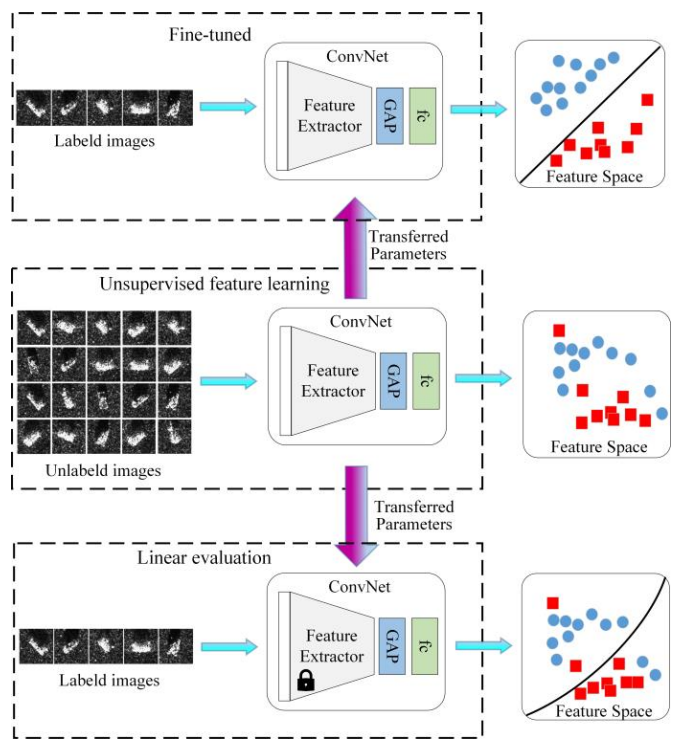

Fig. 6. Illustration of the fine-tuned and the linear evaluation. The lock represents the weights that were frozen. The classifier in the linear evaluation is not limited to the softmax classifier.

Regression [43], Random Forest [44], SVM [45]) without changing the distribution of embedding in the feature space. Fine-tuned and linear evaluation are illustrated vividly in Fig.6.

Undoubtedly, both of them are significant for evaluating self-supervised learning methods. For one thing, linear evaluation examines whether the model has learned features directly related to downstream tasks during the unsupervised feature learning stage. With better performance in linear evaluation, better time-efficiency will be reflected in converging. For another, fine-tuning attends to the potential of the pre-trained model, which directly relates to the performance in downstream tasks. The following experiments results certify the excellent performance of BIDFC in the two evaluations with not only a gratifying convergence speed but also a favorable accuracy.

## IV. EXPERIMENTAL RESULTS AND DISCUSSION

In our experiments, analysis of BIDFC is mainly executed on the MSTAR database while OpenSARShip dataset is also employed to testify the generalization and accuracy of our method. To begin with, we will introduce the procedure of the training set divided into small samples database after the description of MSTAR dataset and OpenSARShip dataset. After that, study of data augmentation and the effect of different hyperparameters of BIDFC on results will be analyzed through ablation study. Then, our method will be compared with the mainstream deep learning technologies in SAR ATR, which are network structure design and transfer learning, followed by the robustness and generalization of BIDFC will be verified. Lastly, comparison between BIDFC and the SOTA methods in other

TABLE I
MSTAR DATASET

| Description | Serial | Training Set | | Serial | Test Set | |
| --- | --- | --- | --- | --- | --- | --- |
| | | Depr | Number | | Depr | Number |
| BMP2 | 9563 | 17° | 233 | 9563 | 15° | 195 |
| | | | | 9566 | 15° | 196 |
| | | | | c21 | 15° | 196 |
| BTR70 | c71 | 17° | 233 | c71 | 15° | 196 |
| T72 | 132 | 17° | 232 | 132 | 15° | 196 |
| | | | | 812 | 15° | 195 |
| | | | | S7 | 15° | 191 |
| ZSU23/4 | D08 | 17° | 299 | D08 | 15° | 274 |
| ZIL131 | E12 | 17° | 299 | E12 | 15° | 274 |
| T62 | A51 | 17° | 299 | A51 | 15° | 273 |
| BTR60 | K10yt7532 | 17° | 256 | K10yt7532 | 15° | 195 |
| D7 | 92v13015 | 17° | 299 | 92v13015 | 15° | 274 |
| BRDM2 | E71 | 17° | 298 | E71 | 15° | 274 |
| 2S1 | B01 | 17° | 299 | B01 | 15° | 274 |
| Total | | | 2747 | | | 3203 |

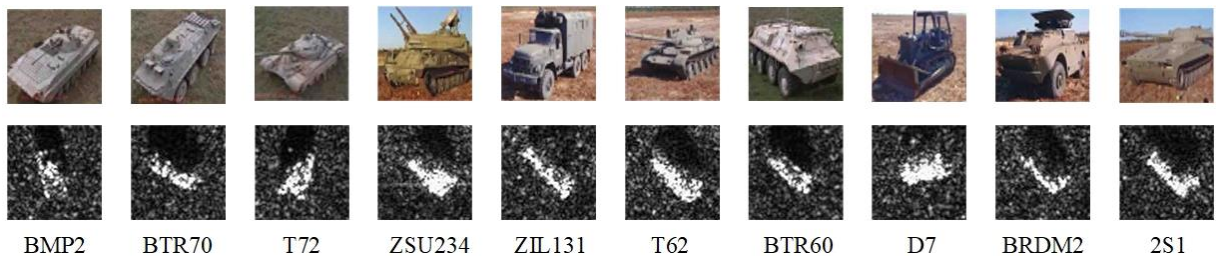

BMP2    BTR70    T72    ZSU234    ZIL131    T62    BTR60    D7    BRDM2    2S1

Fig. 7. SAR Images Instance and Optical Images in MSTAR Database.

computer vision tasks will be made, serving to alleviate the influence of only a small amount of training data (e.g. semi-supervised learning, self-supervised learning, and unsupervised representation learning).

### A. Dataset and Experimental Settings

*1) MSTAR Dataset:* MSTAR [46] is a widely used researched dataset majoring in the SAR targets recognition. Collected by Sandia National Laboratory, this dataset contains ten categories listed in Table I. Each class is sampled from $15°$ or $17°$ depression angle respectively, with corresponding data displayed in Fig.7. Those targets acquired at a depression $15°$ are usually used as testing data, and those at $17°$ as the training data.

*2) OpenSARShip Dataset:* OpenSARShip is presented by Huang *et al.* [47], containing 11346 ship chips from 41 Sentinel-1 SAR images. This database includes 17 type of ships (e.g cargo, fishing, search), but the sample amount of each category scatters extremely uneven. Learning from Huang *et al.* [8], we select a portion of the data from three elaborated types of cargo for the database construction. The training set and testing set are consistent with them. The difference lies in that we take other samples of the three types of ships as unlabeled data to perform BIDFC. This database is applied to verify the generalization of our method, and the details are given in Table II and Fig.8.

*3) Small Samples Database:* The database can be divided into small sample datasets in two ways. The first one is that each category in the training set is sampled proportionally as training data for that category in the small sample database. For example, we can take 1:32 (about 3.13%) of the dataset as the training set. In this way, we create 6 small samples database: 1:32, 1:16, 1:8, 1:4, 1:3, 1:2 of the original training set. The second method is to extract several equivalent samples of each category, naming it as the few-shot database. 1-shot, 2-shot, 3-shot, 4-shot, and 5-shot produces our another experimental database. The former dataset with more samples compared to the second type is titled small sample database. It is worth noting that the above operation can only be conducted on the training set but not the testing set.

*4) Experimental Settings:* For the augmentation of BIDFC, we use crop with resize, random rotation, color jitter, random horizontal flip, Gaussian noise, where the augmentation times K is set to 5. For unsupervised feature learning, ResNet18 is utilized as the ConvNet, and trained at batch size 512 for 300

TABLE II
OPENSARSHIP DATABASE IN OUR EXPERIMENTS

| Elaborated Type | Cargo | Bulk Carrier | Container Ship | Total |
|---|---|---|---|---|
| train | 100 | 100 | 100 | **300** |
| test | 79 | 132 | 135 | **346** |
| unlabeled | - | - | - | **3129** |

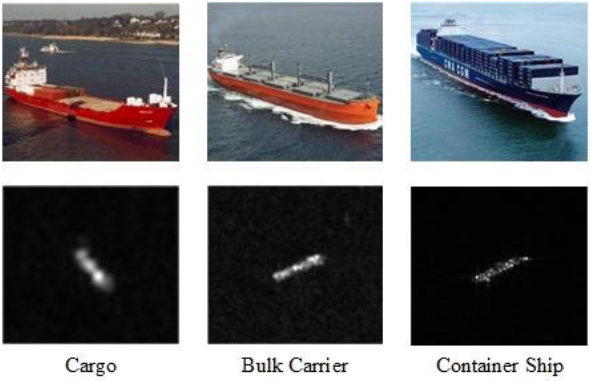

Cargo      Bulk Carrier      Container Ship

Fig. 8. SAR Images Instance and Optical Images in OpenSARShip Database.

epochs. With the use of softmax classifier in fine-tuned or linear evaluation, the model is trained at batch size 256 for 200 epochs. Moreover, Cross-Entropy loss and DWV loss are also incorporated, in which the temperature coefficient T is set to 2.0 in Cross-Entropy loss, while $T_1 = 30$, $T_2 = 150$, and $a_f = 1.0$ are regulated in the weight of DWV Loss, setting the momentum coefficient for parameters update to 0.999. We use Adam as the optimizer, and the initialized learning rate turns out to be linear scaling (i.e. $lr = 0.3 * \text{BatchSize} / 512$) for unsupervised representation learning. In the fine-tuned/linear evaluation, Adam is deployed with initialized learning rate of 0.03. To ensure the reproductivity of the experiment, the seed is arranged to 10. All the experiments are conducted with Intel Core i7-9700K CPU in a Ubuntu 16.04 operation system. The computer is configured with NVIDIA GTX 1080 with 8G RAM where the Pytorch framework is implemented to perform our experiments.

## B. Ablation Studies

In this part, we empirically show the effectiveness of our hyperparameters choice with the following content: augmentation combinations, the iterations of BIDFC, augmentation times $K$ and temperature coefficient T, the values of the momentum coefficient, batch size, learning rate on fine-tuning, the choice of network architecture, classifier for linear evaluation. To enhance the efficiency of the experiment, the logistic regression classifiers for linear evaluation are in place, and a 1:32 set is adopted for the training dataset. Furthermore, the batch size for BIDFC and fine-tuned/linear evaluation in the ablation study is fixed as 256 to ensure that all experiments are available with 8G RAM.

TABLE III
LINEAR EVALUATION ON 1:32 SET OF MSTAR DATABASE UNDER VARIOUS AUGMENTATIONS

| Augmentation | CR | RR | CJ | RHF | GN | GB | Accuracy |
|---|---|---|---|---|---|---|---|
| *One* | ✓ | | | | | | 45.55 |
| | | ✓ | | | | | **58.78** |
| | | | ✓ | | | | 27.56 |
| | | | | ✓ | | | 42.16 |
| | | | | | ✓ | | 42.74 |
| | | | | | | ✓ | 41.61 |
| *Two* | ✓ | | ✓ | | | | 58.41 |
| | ✓ | | | ✓ | | | 58.94 |
| | | | ✓ | ✓ | | | 34.46 |
| | | | ✓ | | ✓ | | **68.24** |
| | ✓ | | | | | ✓ | 55.35 |
| | | | | ✓ | ✓ | | 49.95 |
| *Three* | ✓ | | ✓ | | | | 39.38 |
| | ✓ | ✓ | | ✓ | | | 63.64 |
| | | ✓ | ✓ | | ✓ | | 63.65 |
| | | ✓ | ✓ | | ✓ | | 72.46 |
| | | ✓ | ✓ | | ✓ | | **76.39** |
| | ✓ | | | | | ✓ | 64.02 |
| *Four* | ✓ | ✓ | | | | ✓ | **82.51** |
| | ✓ | | | ✓ | | ✓ | 80.24 |
| | | ✓ | ✓ | ✓ | ✓ | | 41.58 |
| | | ✓ | ✓ | ✓ | ✓ | | 53.48 |
| | ✓ | | ✓ | ✓ | ✓ | | 81.66 |
| | | ✓ | ✓ | | ✓ | ✓ | 78.52 |
| *Five* | ✓ | ✓ | | ✓ | ✓ | ✓ | 84.54 |
| | ✓ | | ✓ | ✓ | ✓ | ✓ | 81.36 |
| | ✓ | ✓ | ✓ | ✓ | | ✓ | 80.89 |
| | ✓ | ✓ | ✓ | ✓ | ✓ | | **88.44** |
| | | ✓ | ✓ | ✓ | ✓ | ✓ | 74.33 |
| | ✓ | | ✓ | ✓ | ✓ | ✓ | 74.48 |
| *Six* | ✓ | ✓ | ✓ | ✓ | ✓ | ✓ | **84.88** |

*1) Augmentation Combinations:* Since the performance of representation learning is readily subject to the influence of the internal parameters of each operation, further descriptions of the parameters of these augmentations methods will be provided below. Crop with resize is equipped with two internal parameters: scale and size. The former one is a range value from $a$ to $b$ capable of controlling the range of size of the original image size cropped, and the $a$ is adjusted to 0.85, while $b$ to1.0. The latter one intends to control the size of the output image, set to 64 consistent with those in other MSTAR images. The degree is the parameter of random rotation operation, set to 30 i.e. the degree range from -30° to 30° will be randomly applied to the original image transformation. Color jitter endows the structure of the image with brightness, contrast, and saturation. Both of brightness and contrast are controlled as 0.4 while the saturation as 0 considering the grayscale of SAR image. Random horizontal flip only needs to consider the probability of this operation which is set to 0.5. Density is related to the strength of the Gaussian noise, thus whose mean is fixed to 0 and variance to 1. Since there is already lots of speckle noise in the SAR images, the density is adjusted to 0.1. The internal parameter of Gaussian blur is the radius (a range value), which is set to [0.1, 2.0]. To promote the efficiency of acknowledging the optimal augmentations, random sampling is employed to combine the six operations instead of using a grid search.

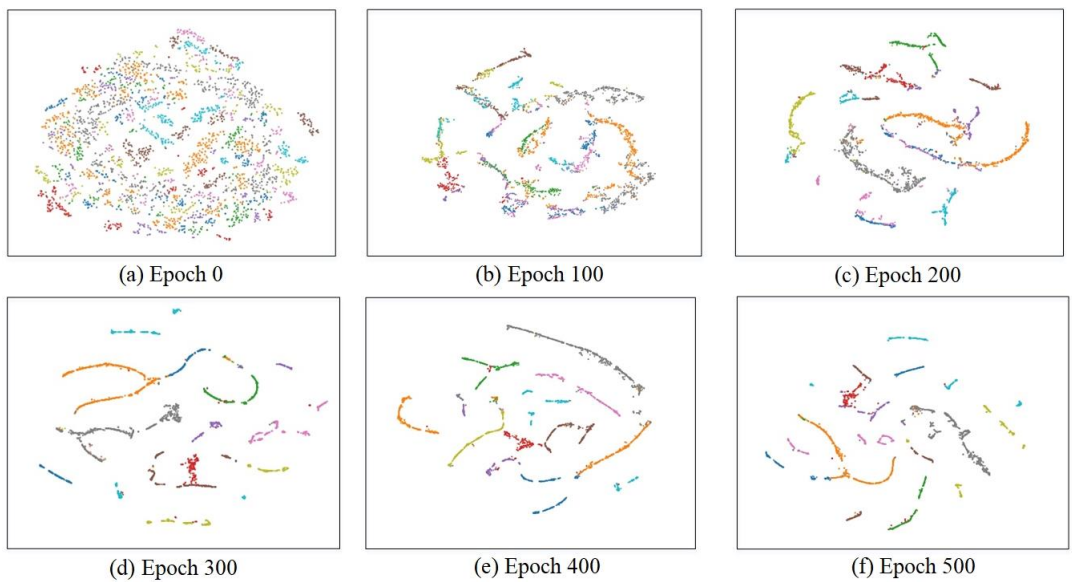

(a) Epoch 0    (b) Epoch 100    (c) Epoch 200

(d) Epoch 300    (e) Epoch 400    (f) Epoch 500

Fig. 9. Visualization of the feature representation from the global average pooling layer of neural network on the MSTAR dataset. Note that the representation visualization ranges from epoch 0 to 500.

Table III evidences that random rotation gives the greatest contributions to SAR images feature learning. Overall, the more complex the augmentation combination is, the better the learned representation is. However, when the six operations are combined, the accuracy of linear evaluation decreases instead. Thus, we decide to take on five data augmentation combination, including RR, CR, CJ, RHF, and GN, to operate contrastive learning.

*2) The Iterations of BIDFC:* We train the ConvNet for 500 epochs via BIDFC, and obtain the embedding of testing data from the global average pooling layer. The high-dimensional embedding is reduced to 2 dimensions via T-sne [48], and visualized with its outcomes in Fig.9. We found that our method learned impressive visual representation, which was reflected in the samples belonging to the same category concentrated together. Besides, after training more than 300 epochs, the compactness of the features did not change significantly. Hence, we set the iterations to 300 epochs.

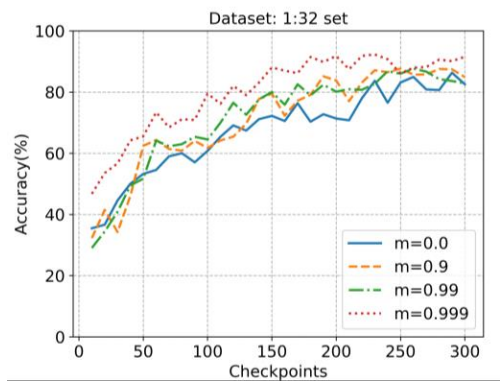

Fig. 10. Linear evaluation on 1:32 set of MSTAR database with different momentum coefficient. The classifier of linear evaluation is logistic regression.

hyperparameters based on the MSTAR dataset with the results listed in Table IV.

Information from Table IV tells that when $T$=2.0 and $K$=7, the accuracy of the linear evaluation outperforms all the other ones, but the data of $K$=5 and $K$=7 when $T$=2.0 differ a few. While $K$ reaches 7, more memory and longer training time are required so we set the $T$ to 2.0 and $K$ to 5 for the optimal operation. Besides, the experiment testifies that extreme $T$ will lead to awfully weak intensity of contrastive learning, which may significantly disable the model to converge or impair the representations learned from BIDFC. As $K$ increases while $T$ decreases, contrastive learning is strengthened while the performance of the ConvNet begins to decline correspondingly.

*4) Momentum Coefficient:* The parameter can stabilize the process of BIDFC, and further improve the learned features. So we perform the linear evaluation on 1:32 set with different values, which uses checkpoints learned on the unsupervised feature learning stage. The results are depicted in Fig 10, acknowledging the best figure for m in unsupervised feature learning turns out to be 0.999.

TABLE IV
LINEAR EVALUATION WITH DIFFERENT $K$ AND $T$ UNDER
1:32 SET OF MSTAR DATABASE

| $K$ | 1 | 3 | 5 | 7 | 9 | 11 |
|---|---|---|---|---|---|---|
| $T$=0.5 | *fail* | *fail* | 74.96 | 85.12 | 82.16 | 64.40 |
| $T$=1.0 | *fail* | *fail* | 85.26 | 84.48 | 81.26 | 79.64 |
| $T$=1.5 | *fail* | *fail* | 87.69 | 88.12 | 78.55 | 73.99 |
| $T$=2.0 | *fail* | *fail* | 88.13 | **88.44** | 82.98 | 77.89 |

*3) Augmentation Times $K$ and Temperature Coefficient $T$:* Augmentation times and temperature coefficient controls the degree of "weak" in the weakly contrastive learning. Here we experiment with different $K$ and $T$ to find the optimal

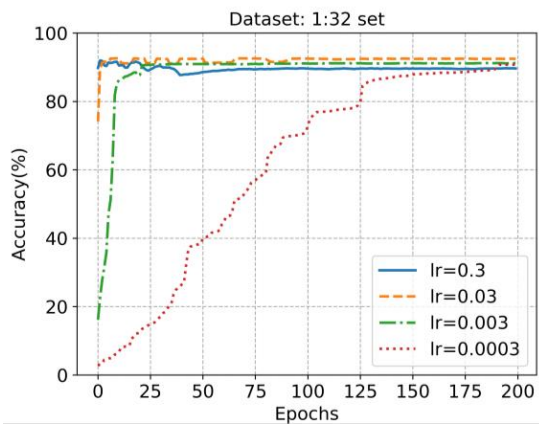

Fig. 11. Fine-tuned on 1:32 set of MSTAR with different learning rates.

*5) Batch Size:* Table V shows the impact of batch size for BIDFC. We train different batch size on 300 epochs, and linear evaluate the representation learned by BIDFC. Experimental data illustrates that batch size is positively related to the accuracy. What is more, it is acknowledged that larger batch size increases the number of instances for identification, which can raise the complexity of pretext tasks, softens the probability distribution and reduces the intensity of contrastive learning. Since NVIDIA GTX 1080 with 8G RAM is used in the paper, experiments with a batch size higher than 512 cannot be performed due to memory overflow.

TABLE V
LINEAR EVALUATION WITH DIFFERENT BATCH SIZE UNDER
1:32 SET OF MSTAR DATABASE

| Batch Size | 32 | 64 | 128 | 256 | 512 |
|---|---|---|---|---|---|
| Accuracy (%) | *fail* | 81.67 | 84.67 | 88.44 | **90.13** |

*6) Learning rate on fine-tuned stage:* Varied from fine-tuned on the pre-trained parameters (e.g. 3x10^-3, 3x10^-4), our proposed method can be exploited even with a larger learning rate, resulting in faster model converge and higher recognition rate. Fig.11 proves that when a larger learning rate adopted, the model can still converge quickly and realize a high recognition rate through our method. Comparison between data defines 0.03 as our optimal learning rate.

*7) Network Architecture:* The choice of network architecture has a significant impact on unsupervised feature learning. The results of [21] show that great depth of the network can help generate better learned features. Therefore, discussion for the impact of using ResNet18, ResNet34, ResNet50 on feature representation has been raised. Since the channels in the last convolution layer of ResNet50 have 2048 dimensions, repetitive feature clustering in BIDFC certainly induces a huge computation cost. Therefore, we replace the basic block to the bottleneck in ResNet50, and linear evaluate the representation learned by these networks, with the *K* set to 7 when training ResNet34 and ResNet50 to help the model converge better.

TABLE VI
LINEAR EVALUATION WITH DIFFERENT ARCHITECTURE
UNDER 1:32 SET OF MSTAR DATABASE

| Architecture | ResNet18 | ResNet34 | ResNet50 |
|---|---|---|---|
| Accuracy (%) | **88.44** | 85.73 | 81.79 |

The results are shown in Table VI. Nevertheless, divergence has been spotted between our results and Chen's conclusion [16]. Our findings inform us of the fact that the lightweight network can learn better representation in SAR images. It is believed that scarce amount of data in the MSTAR database can lead to overfitting during the unsupervised feature learning stage, so ResNet18 with smaller parameters will be the choice for the following experiments.

*8) Classifiers for Linear Evaluation:* In this section, performances of different classifiers in linear evaluation are scrutinized. The experimental results in Table VII report that our method is also accompanied by considerable accuracy under the linear evaluation, which relates the benefits of the features learned in the unsupervised feature learning stage to the downstream tasks. In short, through comparison, the softmax classifier possesses the highest recognition rate under most of the situation even though it is time-consuming, while logistic regression seems to be the optimum with acceptable accuracy and speed.

TABLE VII
LINEAR EVALUATION ON SMALL SAMPLE MSTAR DATABASE
WITH DIFFERENT CLASSIFIERS

| Method | 1:32 | 1:16 | 1:8 | 1:4 | 1:3 | 1:2 |
|---|---|---|---|---|---|---|
| Softmax | 90.06 | **91.79** | **93.72** | **93.41** | **94.02** | **94.10** |
| KNN | 74.89 | 86.25 | 91.66 | 92.06 | 92.13 | 92.56 |
| SVM | 87.66 | 91.57 | 91.75 | 91.93 | 91.94 | 92.28 |
| Random Forest | 89.23 | 90.22 | 91.78 | 91.97 | 93.22 | 92.03 |
| Logistic Regression | **90.13** | 91.53 | 91.94 | 92.38 | 92.35 | 92.35 |

### C. Comparison with Classic/Specific Network and Transfer Learning

Researchers normally identify SAR images with the classic network or specific model designed by themselves and further improve the recognition rate via transfer learning. In this part, two comparisons will be developed: one with classic/specific network trained from scratch; the other with transfer learning the ImageNet, fine-tuned on the representation learned by BIDFC.

*1) Comparison with Classic/Specific Network:* Zhai *et al.* [49] proposed a specific network MFFA-SARNET, compared with different classic network based on small sample datasets which included LeNet [26], AlexNet [27], VGG16 [28], ResNet50 [29], Inception V3 [30], DenseNet [31], and Xecption [32]. H_Net [6] and A_ConvNet [7] are applied in small sample learning in this paper. Records from comparisons of fine-tuning with different networks based on the small sample database are detailed in Table VIII. Results account that

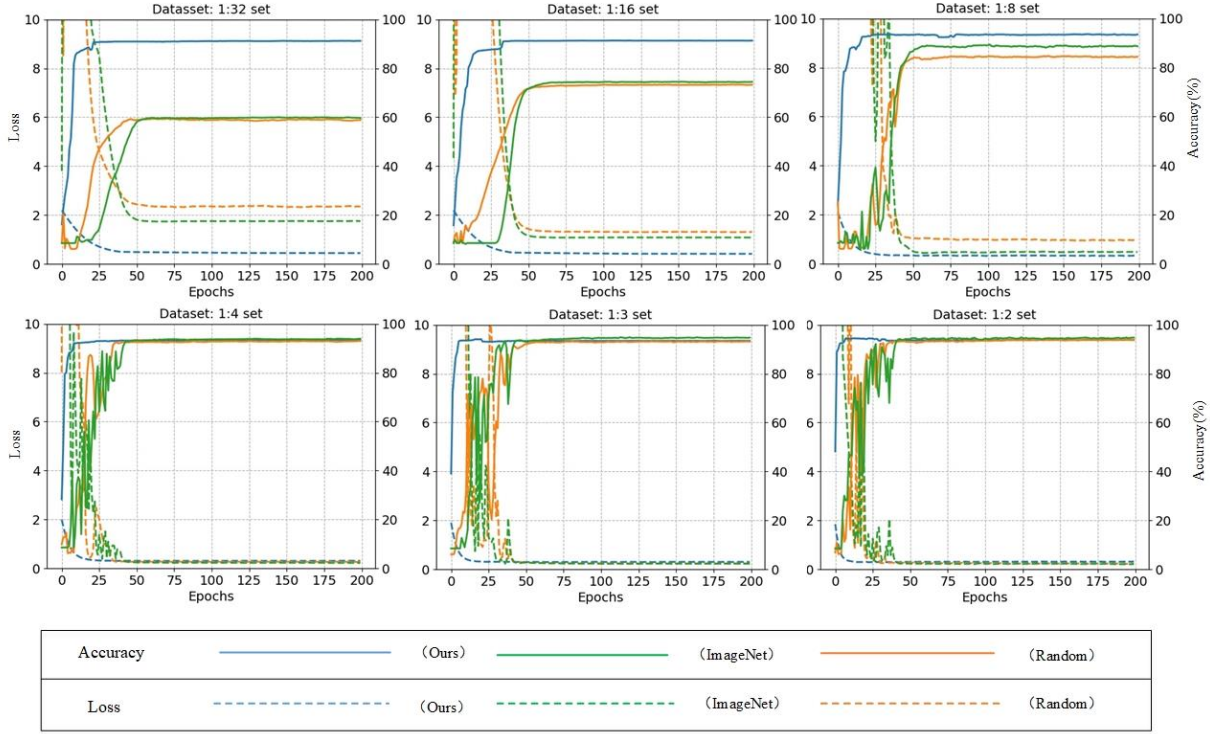

Fig. 12. The ResNet18 utilizes three different initial weights to fined-tune on small sample datasets. The solid line represents the accuracy of the testing set while the dashed line represents the testing loss. Ours: fine-tuned on the representation learned from the BIDFC. ImageNet: fine-tuned on the model pre-trained on the ImageNet database. Random: train from scratch.

TABLE VIII

Comparasion with Classic Network on Small Sample MSTAR Database

| Method | 1:32 | 1:16 | 1:8 | 1:4 | 1:3 | 1:2 |
|---|---|---|---|---|---|---|
| LeNet | 31.60 | 48.00 | 63.00 | 80.90 | 84.10 | 87.40 |
| AlexNet | 38.20 | 60.50 | 80.90 | 89.70 | 92.40 | 93.20 |
| VGG16 | 14.00 | 31.40 | 39.20 | 46.10 | 55.50 | 62.40 |
| ResNet50 | 20.10 | 25.10 | 28.50 | 35.60 | 59.70 | 69.00 |
| InceptionV3 | 17.70 | 43.50 | 49.80 | 60.20 | 71.10 | 82.00 |
| DenseNet | 17.00 | 50.00 | 60.80 | 82.00 | 87.50 | 84.20 |
| Xecption | 22.50 | 34.10 | 41.60 | 70.40 | 78.50 | 84.00 |
| MFFA-SARNET | 38.00 | 51.80 | 69.50 | 92.90 | 95.40 | 96.60 |
| A_ConvNet | 63.54 | 72.12 | 76.80 | 88.73 | 87.23 | 94.38 |
| H_Net | 52.33 | 61.47 | 71.42 | 84.38 | 89.45 | 91.13 |
| BIDFC (ours) | 91.25 | 92.34 | 95.22 | 96.25 | 96.48 | 97.26 |

TABLE IX

Fine-tuned on BIDFC, Imagenet and Random Initialization on Small Sample MSTAR Database

| Method | 1:32 | 1:16 | 1:8 | 1:4 | 1:3 | 1:2 |
|---|---|---|---|---|---|---|
| ImageNet | 59.26 | 74.52 | 89.35 | 92.49 | **94.87** | 94.87 |
| Random | 60.47 | 73.39 | 84.67 | 91.68 | 93.28 | 93.94 |
| BIDFC(Ours) | **90.12** | **91.35** | **93.63** | **93.76** | 94.09 | **95.50** |

the model structure cannot prominently improve the performance of small sample training dataset, but common model with BIDFC shows the opposite. To conclude, the model structure for small sample SAR ATR does not play a decisive role in its performance.

*2) Comparison with Transfer Learning:* Transfer learning as a common technique in the field of SAR ATR, employs the ImageNet pre-trained model to fine-tuned. Many scholars also have explored various new parameters transferred methods such as those with specific layers frozen or novel objective functions for domain adaption. Furthermore, these tricks are also applicable to our fine-tuned stage. For fair comparison and highlight on the excellent features we learned in the unsupervised feature learning stage, we fined-tuned the feature learned by BIDFC to compare fined-tuned with the ImageNet pre-trained model and trained from the scratch in the case of same architecture and training set. To ensure better convergence for the other two methods, we here apply a smaller learning rate 0.003 for three methods of fine-tuning. Circumstantial results can be referred to Table IX and Fig.12.

The results notify that our method exceeds Random Initialization and ImageNet pre-trained in the case of small amounts of data in terms of convergence speed and accuracy. As the number of training data rises, the accuracy of these methods begins to approach. When using more than 1:3 training data, fine-tuned on the ImageNet weights even outperforms our method. Although this is not the optimal performance of our method, our method still owns the best suitability if only a small amount of training data provided. Moreover, since the gap between fine-tuned on the weights of Random and IamgeNet tends to be minute, it is not a good choice to apply ImageNet pre-trained parameters to non-optical images training.

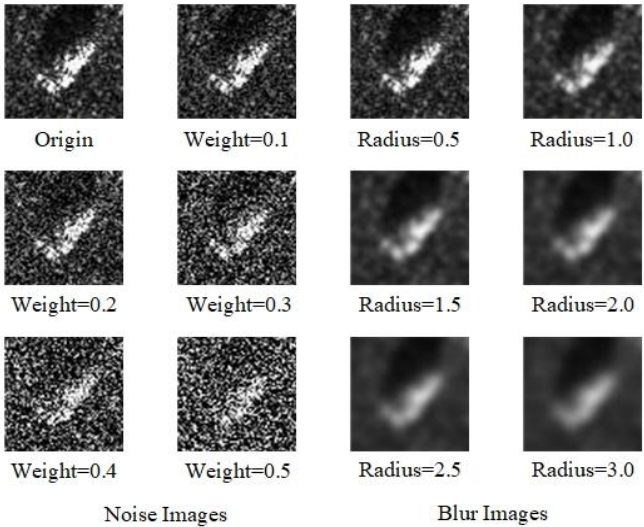

Noise Images　　　　　　　　Blur Images

Fig. 13. The first and second columns are noise images with various degrees of disturbance. The third and fourth columns are different degrees of blurred images.

### D. Bigger Challenge

In this part, the superiority of our method will be further examined. First, noise and fuzzy disturbance will be artificially added to the image to verify the adaption of the model. Then the model will be fine-tuned/linear evaluated on a smaller database (i.e. few-shot dataset). In addition, the pre-trained model learned by BIDFC fine-tuned on the noisy label data will be studied to check the robustness of the representations. Ultimately, our method will be applied to other SAR datasets in order to inspect the generalization of our method.

*1) Blur and Noise:* As an inherent issue in the SAR images generation system, blur and noise can lower the image quality and damage the SAR ATR. To attest the minial influence of blur and noise on BIDFC, various degrees of noise and blur perturbation are enforced on the small sample dataset respectively. For Image noise, we apply Gaussian noise whose means as 0 and variance as 1, with various weights from 0.1 to 0.5 to adjust the noise intensity while applying Gaussian blur to images that the radius varied from 0.5 to 3.0. Images affected by blur and noise are enumerated in Fig.13.

Through utilizing the pre-trained model via BIDFC to launch contrastive learning on original training dataset and to perform linear evaluation and fine-tuning on small sample database with noise and blur disturbance, it is believed that our method can overcome the perturbation and simultaneously maintain a considerable recognition rate. The results are expounded in Fig.14 and Fig.15.

*2) Smaller Datasets:* In the experiments above, comparison is mainly made between the small sample dataset and other methods, but attention should also be cast to BIDFC's performance on smaller datasets. Thus, the results of fined-tuned/linear evaluation on the few-shot database are organized in Table X, which again confirm high accuracy of our method in few-shot learning.

*3) Noisy Label:* The robustness of the model can prominently impact its practical application. One of the most intriguing phenomena for deep learning belongs to that a small amount of noisy data can greatly reduce the performance of the model [50]. Therefore, we introduce noisy label to small sample datasets and check the performance of the model training on this sample. The way adopted to create noise datasets is to introduce other class of data in a certain portion for each category. For example, in the 1:32 set, BMP2 contains 7 samples. We create a noise database with a ratio of 0.3, and now there are 5 correct samples and 2 noise data. We create noise database with noise ratio of 0.3, 0.5 and 0.7 for small sample datasets (i.e. 1:32, 1:16, 1:8, 1:4, 1:3, 1:2), respectively. The model will be fine-tuned and linear evaluated using logistic regression on these datasets (In Fig.16). Because the noise data may destabilize the model convergence, we will record the stable value instead of the highest value.

Our method can efficiently resist the influence of noisy data. Normally, a small amount of training samples can apparently impact the performance. However, unlike the previous experiments, the performance of linear evaluation defeats that of the fine-tuned in most cases, and even outstrips the noise-free data in some datasets (e.g. 1:3, 1:2 database). With these results, we approximate that freezing the weights of feature extractors can effectively avoid the damages of the noisy label to the feature representation.

*4) Generalization:* To verify the generalization of our method, we apply BIDFC to the unlabeled data in our self-built OpenSARShip dataset and compare our fine-tuning results with Huang *et al.* [8]. Three network architectures were applied in their paper, namely H_Net [6], A_ConvNet [7], and AlexNet [27] and the influence of different source domains for transfer learning was discussed, including MSTAR [46], ImageNet [24], and SAR (reconstruct). Table XI shows their results and the BIDFC based on different networks and source domains.

### TABLE X
### Fine-tuned/Linear Evaluation on Few-shot MSTAR Datasets

| Method | 1-shot | 2-shot | 3-shot | 4-shot | 5-shot |
|---|---|---|---|---|---|
| *Fine-tuned:* | | | | | |
| BIDFC (ours) | 80.74 | 85.30 | 87.26 | 88.42 | 90.26 |
| *Linear evaluation:* | | | | | |
| Softmax | **75.93** | 82.24 | 82.05 | **87.70** | **88.62** |
| KNN | 73.68 | **84.51** | **86.54** | 87.07 | 88.47 |
| SVM | 73.68 | 83.57 | 82.36 | 84.88 | 87.76 |
| Random Forest | 74.14 | 82.67 | 80.89 | 83.01 | 86.41 |
| Logistic Regression | 72.67 | 82.29 | 85.49 | 85.95 | 87.91 |

### TABLE XI
### OpenSARShip Recognition Rate on Different Networks and Source Tasks

| Network | Source Task | Random Initialization | Fine-tuned |
|---|---|---|---|
| **A_ConvNet** | MSTAR | 87.57 | 86.70 |
| **H_Net** | SAR (reconstruct) | 85.55 | 86.41 |
| | MSTAR | | 88.05 |
| **AlexNet** | ImageNet | 84.39 | 89.01 |
| | SAR | | 89.21 |
| | MSTAR | | 90.75 |
| **ResNet18** | ImageNet | 82.02 | 85.37 |
| | OpenSARShip (unlabeled) | | **91.91** |

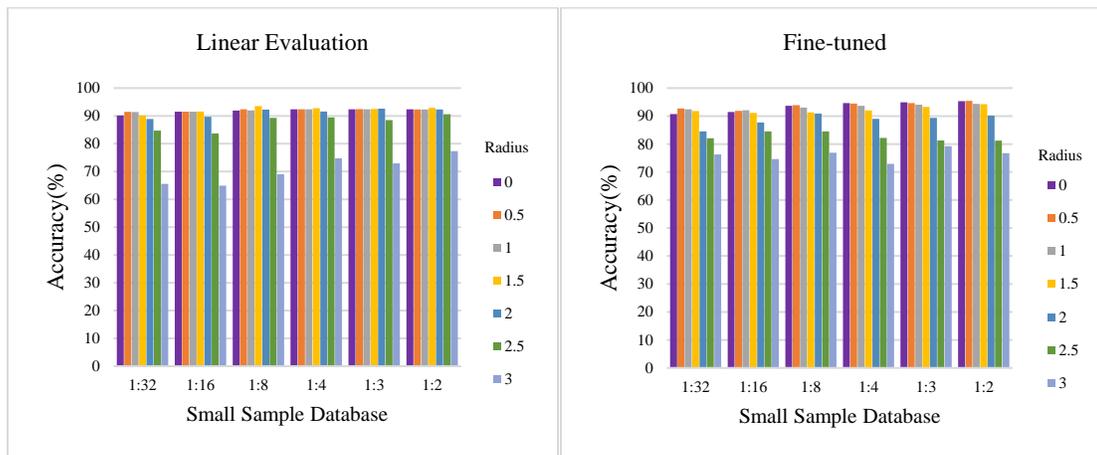

Fig. 14. Fine-tuned/Linear evaluation on small sample database with various fuzzy disturbance.

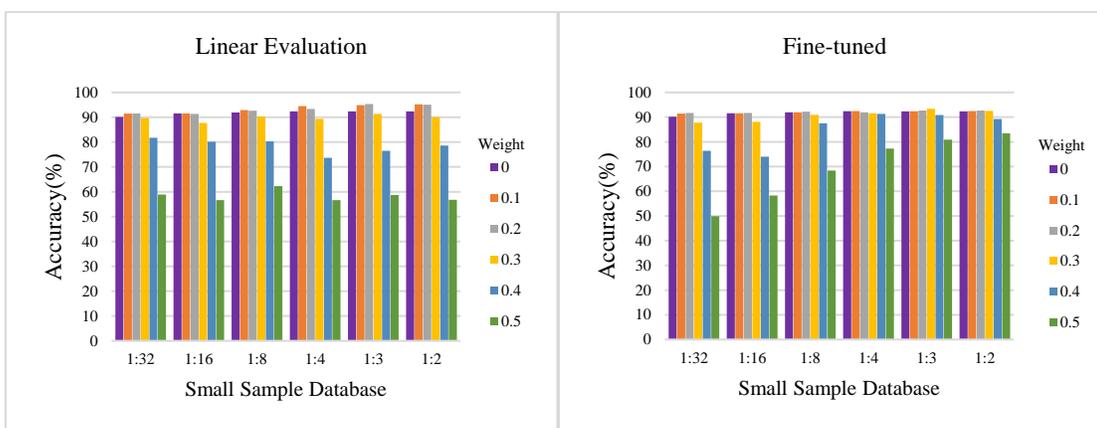

Fig. 15. Fine-tuned/Linear evaluation on small sample database with various noise disturbance.

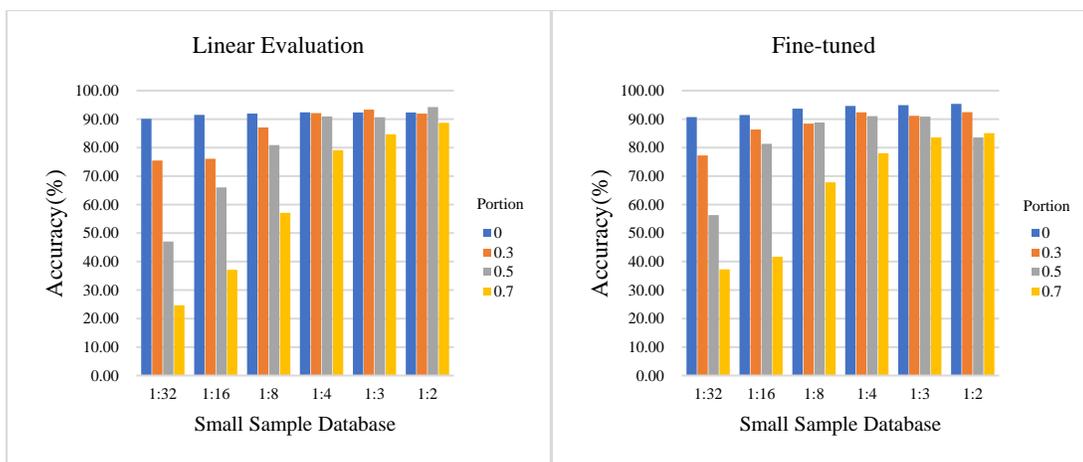

Fig. 16. Fine-tuned/Linear evaluation of different small sample noisy label datasets.

### E. Comparison with SOTA Methods

In this section, we will compare our proposed method with the SOTA methods to counter the small sample problems, including semi-supervised learning, self-supervised learning, and unsupervised representation learning. We train semi-supervised methods with Pseduo-Label [40] and VAT [51], which deploy small sample database for supervised learning, and the rest of the samples as unlabeled data for unsupervised learning. In self-supervised learning, comparisons are conducted among Rotation [16], InstDisc [20], Simclr [21], Moco [22]. In terms of unsupervised representation learning, VAE [52] will be evaluated on SAR ATR. Since the semi-supervised learning has no feature learning stage, we do not perform the linear evaluation on VAT and Pseudo-Label. Besides, these methods will be trained on ResNet18 for 300 epochs. Simclr, Moco, and Instdisc will be trained with the augmentations proposed in this paper for the unsupervised representation learning stage. Table XII provides data from different methods for the convenience of our comparison. Based on the SAR ATR, the experimental results show that BIDFC is better than strongly contrastive learning methods i.e. InstDisc, Simclr, and Moco, which is more conducive to learning useful semantic features and verifies the correctness of our conjecture.

TABLE XII
SOTA METHODS TRAINED ON SMALL SAMPLE MSTAR DATABASE

| Method | 1:32 | 1:16 | 1:8 | 1:4 | 1:3 | 1:2 |
|---|---|---|---|---|---|---|
| *Fine-tuned:* | | | | | | |
| Pseduo-Label | 65.35 | 82.51 | 86.69 | 92.25 | 93.47 | 94.22 |
| VAT | 70.37 | 82.79 | 93.31 | **96.62** | 96.03 | 95.62 |
| Rotataion | 66.75 | 76.08 | 87.03 | 90.54 | 93.23 | 93.41 |
| InstDisc | 67.62 | 82.20 | 89.85 | 93.06 | 93.81 | 94.22 |
| Simclr | 84.86 | 88.04 | 93.19 | 93.88 | 94.54 | 94.94 |
| Moco | 81.27 | 86.51 | 93.23 | 94.15 | 94.22 | 94.31 |
| VAE | 53.29 | 71.99 | 88.13 | 92.81 | 92.44 | 94.84 |
| BIDFC(ours) | **91.25** | **92.34** | **95.22** | 96.25 | **96.48** | **97.26** |
| *Linear evaluation :* | | | | | | |
| Pseduo-Label | - | - | - | - | - | - |
| VAT | - | - | - | - | - | - |
| Rotation | 41.61 | 51.73 | 56.89 | 58.81 | 60.09 | 62.34 |
| InstDisc | 57.07 | 64.28 | 75.77 | 82.39 | 84.42 | 87.57 |
| Simclr | 84.38 | 87.88 | 91.66 | 91.50 | 91.78 | 91.41 |
| Moco | 79.46 | 82.37 | 86.59 | 87.17 | 88.47 | 88.69 |
| VAE | 27.53 | 30.81 | 41.71 | 47.64 | 51.60 | 54.91 |
| BIDFC(ours) | **90.13** | **91.53** | **91.94** | **92.38** | **92.35** | **92.35** |

## V. CONCLUSION

In this paper, motivated by contrastive learning and the characteristics of SAR images, we propose a weakly contrastive learning method, termed as Batch Instance Discrimination and Feature Clustering (BIDFC). This method addresses the issue of strongly contrastive learning damaging the semantic features learned based on SAR images, via controlling the intensity of contrastive learning. Through massive experiments, we validate the superiority of our method based on the MSTAR and OpenSARShip benchmark, and its obvious edge in sustaining accuracy with only a small amount of training data. Besides, due to the highly hierarchical image features learned by our method, the model has strong adaptability even under fuzzy disturbance, noisy disturbance and noisy label. This is of great significance for the deployment of the SAR ATR system in actual scenarios. Consequently, it is expected that our framework can be widely adopted in the field of SAR ATR to obtain a more robust, and cost-efficient model. In the future, further endeavors will be invested to unsupervised learning in hope of receiving a considerable recognition rate without any annotated data.

## REFERENCES

[1] W. M. Brown, "Synthetic Aperture Radar," *IEEE Trans. Aerosp. Electron. Syst.*, vol. AES-3, no. 2, pp. 217-229, 1967.

[2] R. K. Raney *et al.*, "Precision SAR processing using chirp scaling," *IEEE Trans. Geosci. Remote Sens.*, vol. 32, no. 4, pp. 786-799, Mar. 1994.

[3] C. Clemente *et al.*, "Automatic Target Recognition of Military Vehicles with Krawtchouk Moments," *IEEE Trans. Aerosp. Electron. Syst.*, vol. 53, no.1, pp. 493-500, 2017.

[4] Y. Sun, L. Du, Y. Wang, and J. Hu, "SAR Automatic Target Recognition Based on Dictionary Learning and Joint Dynamic Sparse Representation," *IEEE Geosci. Remote Sens.*, vol. 13, no. 99, pp. 1777-1781, 2016.

[5] C. Clemente *et al.*, "Pseudo-Zernike Based Multi-Pass Automatic Target Recognition From Multi-Channel SAR," *IET Radar, Sonar & Navigation*, vol. 9, no. 4, pp. 457-466, 2015.

[6] Z. Huang, Z. Pan, and B. Lei, "Transfer learning with deep convolutional neural network for SAR target classification with limited labeled data," *Remote Sens.*, vol. 9, no. 9, pp. 907-928, Aug. 2017.

[7] S. Chen, H. Wang, F. Xu, and Y. Q. Jin, "Target classification using the deep convolutional networks for SAR images," *IEEE Trans. Geosci. Remote Sens.*, vol. 54, no. 8, pp. 4806–4817, Aug. 2016.

[8] Z. Huang, Z. Pan, and B. Lei, "What, Where, and How to Transfer in SAR Target Recognition Based on Deep CNNs," *IEEE Trans. Geosci. Remote Sens.*, vol. 58, no. 4, pp. 2324-2336, Oct. 2019.

[9] Z. Chen, T. Zhang, and C. Ouyang, "End-to-End Airplane Detection Using Transfer Learning in Remote Sensing Images," *Remote Sens.*,vol. 10, no. 1, pp. 139-153, Jan. 2018.

[10] L. Schmarje, M. Santarossa, S. Schröder, and R. Koch, "A survey on Semi-, Self- and Unsupervised Techniques in Image Classification," 2020, *arXiv:2002.08721*. [Online]. Available: https://arxiv.org/abs/2002.08721.

[11] L. Jing *et al.*, "Self-supervised Visual Feature Learning with Deep Neural Networks: A Survey," 2020, *arXiv:1902.06162*. [Online]. Available: https://arxiv.org/abs/1902.06162.

[12] P. Goyal, D. Mahajan, A. Gupta, and I. Misra, "Scaling and Benchmarking Self-Supervised Visual Representation Learning," *in Proc. IEEE Int. Conf. Comput. Vis. (ICCV)*, Seoul, Korea, Mar. 2019, 6391-6400.

[13] T. Chen *et al*, "Adversarial Robustness: From Self-Supervised Pre-Training to Fine-Tuning," *in Proc. IEEE Conf. Comput. Vis. Pattern Recognit. (CVPR)*, Jun. 2020, 696-705.

[14] C. Doersch, A. Gupta, and A. A. Efros, "Unsupervised Visual Representation Learning by Context Prediction," *in Proc. IEEE Int. Conf. Comput. Vis. (ICCV)*, Santiago, Chile, Apr. 2015, pp. 1422-1430.

[15] M. Noroozi and P. Favaro, " Unsupervised Learning of Visual Representations by Solving Jigsaw Puzzles," *in Proc. Eur. Conf. Comput. Vis. (ECCV)*, Amsterdam, Netherlands, Oct. 2016, pp. 69-84.

[16] S. Gidaris, P. Singh, and N. Komodakis, "Unsupervised Representation Learning by Predicting Image Rotations," *in Proc. Int. Conf. Learn. Representations (ICLR)*, Vancouver Canada, May. 2018, pp. 1-16.

[17] L. Beyer, X. Zhai, A. Oliver, and A. Kolesnikov, "S4L: Self-Supervised Semi-Supervised Learning," *in Proc. IEEE Int. Conf. Comput. Vis. (ICCV)*, Seoul, Korea, Mar. 2019, pp. 1476-1485.

[18] R. Zhang, P. Isola, and A. A. Efros, "Colorful Image Colorization," in *Proc. Eur. Conf. Comput. Vis. (ECCV)*, Amsterdam, Netherlands, Oct. 2016, pp. 649-666.

[19] R. Hadsell, S. Chopra, and Y. Lecun, "Dimensionality reduction by learning an invariant mapping," *in Proc. IEEE Conf. Comput. Vis. Pattern Recognit. (CVPR)*, New York, USA, Jun. 2006, pp. 1735-1742.

[20] Z. Wu, Y. Xiong, S. Yu, and D. Lin, "Unsupervised feature learning via non-parametric instance discrimination," *in Proc. IEEE Conf. Comput. Vis. Pattern Recognit. (CVPR)*, Salt Lake City, USA, Jun. 2018, pp. 3733-3742.

[21] T. Chen *et al.*, "A Simple Framework for Contrastive Learning of Visual Representations," 2020, *arXiv:2002.05709*. [Online]. Available: https://arxiv.org/abs/ 2002 .05709.

[22] K. He *et al.*, "Momentum Contrast for Unsupervised Visual Representation Learning," in *Proc. IEEE Conf. Comput. Vis. Pattern Recognit. (CVPR)*, Jun. 2020, pp. 9729-9738.

[23] X. Chen, H. Fan, R. Girshick, and K. He, "Improved Baselines with Momentum Contrastive Learning," 2020, *arXiv:2003.04297*. [Online]. Available: https://arxiv.org/abs/2003.04297.

[24] J. Deng *et al*, "Imagenet: A large-scale hierarchical image sdatabase," in *Proc. IEEE Conf. Comput. Vis. Pattern Recognit. (CVPR)*, Miami, USA, Jun. 2009, pp. 20-25.

[25] D. Morgan, "Deep convolutional neural networks for ATR from SAR imagery," *in proc. SPIE, Algorithms for Synthetic Aperture Radar Imagery XXII*, Orlando, USA, Jul. 2015, pp. 94750F-1 - 94750F-13.

[26] Y. Lecun, L. Bottou, Y. Bengio, and P. Haffner, "Gradient-based Learning Applied to Document Recognition,"*Proceedings of the IEEE*, vol. 86, no. 11, pp. 2278-2324, Apr. 1998.

[27] A. Krizhevsky, I. Sutskever, and G. E. Hinton, "Imagenet classification with deep convolutional neural networks," in *Proc. Adv. Neural Inf. Process. Syst. (NIPS)*, New York, USA, Dec, 2012, pp. 1097–1105.

[28] K. Simonyan and A. Zisserman, "Very Deep Convolutional Networks for Large-scale Image Recognition," in *Proc. Int. Conf. Learn. Representations (ICLR)*, San Diego, May. 2015, pp. 1-14.

[29] K. He, X. Zhang, S. Ren, and J. Sun, "Deep Residual Learning for Image Recognition," in *Proc. IEEE Conf. Comput. Vis. Pattern Recognit. (CVPR)*, Las Vegas, USA, Jun. 2016, pp. 770-778.

[30] C. Szegedy *et al.*, "Rethinking the Inception Architecture for Computer Vision," *in Proc. IEEE Conf. Comput. Vis. Pattern Recognit. (CVPR)*, Las Vegas, USA, Jun. 2016, pp. 2818-2826.

[31] G. Huang *et al.*, "Densely Connected Convolutional Networks," *in Proc. IEEE Conf. Comput. Vis. Pattern Recognit. (CVPR)*, Hawaii, USA, Jul. 2017, pp. 2261-2269.

[32] F. Chollet, "Xception: Deep Learning with Depthwise Separable Convolution," *in Proc. IEEE Conf. Comput. Vis. Pattern Recognit. (CVPR)*, Hawaii, USA, Jul. 2017, pp. 1800-1807.

[33] N. Srivastava, G. E. Hinton, A. Krizhevsky, I. Sutskever, and R. Salakhutdinov, "Dropout: A Simple Way to Prevent Neural Networks from Overfitting," *J. Mach. Learn. Res.*, vol. 15,  no. 1, pp. 1929-1958, 2014.

[34] S. Ioffe and C. Szegedy, "Batch Normalization: Accelerating Deep Network Training by Reducing Internal Covariate Shift," in *Proc. 32nd Int. Conf. Mach. Learn. (ICML)*, Lille, France, Jul. 2015, pp. 448-456.

[35] F. Gao, T. Huang, J. Sun, J. Wang, A. Hussain, and E. Yang, "A New Algorithm of SAR Image Target Recognition Based on Improved Deep Convolutional Neural Network," *Cognitive Computation*, vol. 11, no. 6, pp. 809-824, 2019.

[36] Z. Yue *et al.*, "A Novel Semi-Supervised Convolutional Neural Network Method for Synthetic Aperture Radar Image Recognition," *Cognitive Computation (2019)*. [Online]. Available:  https://doi.org/10.1007/s12559-019-09639-x

[37] Z. Lin, K. Ji, M. Kang, X. Leng, and H. Zou, " Deep Convolutional Highway Unit Network for SAR Target Classification with Limited Labeled Training Data," *IEEE Geosci. Remote Sens.*, vol. 14, no. 7, pp. 1091-1095, Apr. 2017

[38] R. K. Srivastava, K. Greff, and J. Schmidhuber, "Training Very Deep Networks," in *Proc. Adv. Neural Inf. Process. Syst. (NIPS)*, Montreal, Canada, Dec. 2015, pp. 2377-2385.

[39] G. E. Hinton, O. Vinyals, and J. Dean, "Distilling the Knowledge in a Neural Network," in *Proc. Adv. Neural Inf. Process. Syst. (NIPS)*, Montreal, Canada, Dec. 2015, pp. 1-9.

[40] D. Lee, "Pseudo-Label: The Simple and Efficient Semi-Supervised Learning Method for Deep Neural Networks," in *Proc. 30nd Int. Conf. Mach. Learn. Workshop*, Atlanta, USA, Jun. 2013, pp. 1-6.

[41] Q. Xie, Z. Dai, E. Hovy, M. Luong, and Q. V. Le, "Unsupervised Data Augmentation for Consistency Training,". in *Proc. Int. Conf. Learn. Representations (ICLR)*, Apr. 2020, pp. 1-18.

[42] Y. Zhang , J. Wu , J. Wang, and C. Xing, "A Transformation-based Framework for KNN Set Similarity Search," *IEEE Trans. Knowl. Data Eng.*, vol. 32, no. 3, pp. 409-423, 2020

[43] M. Tan, I. W. Tsang, and L. Wang, "Minimax Sparse Logistic Regression for Very High- Dimensional Feature Selection," *IEEE Trans. Nerual Netw. Learn. Syst.*, vol. 24, no. 10, pp.  1609-1622, May. 2013.

[44] L. Breiman, "Random Forests," *Mach. Learn.*, vol. 45, no. 1, pp. 5-32, Oct. 2001.

[45] C. Cortes and V. Vapnik, "Support-Vector Networks," *Mach. Learn.*, vol. 20, no. 3, pp. 273-297, Feb. 1995.

[46] *The Air Force Moving and Stationary Target Recognition Database*. Accessed: Feb. 3, 2016. [Online]. Available:  https://www.sdms.afrl.af.mil/datasets/mstar/

[47] L. Huang *et al.*, "OpenSARShip: A dataset dedicated to Sentinel-1 ship interpretation," *IEEE J. Sel. Topics Appl. Earth Observ. Remote Sens.*, vol. 11, no. 1, pp. 195–208, Jan. 2018

[48] L. V. D. Maaten and G. E. Hinton, "Visualizing data using t-SNE," *J. Mach. Learn. Res.*, vol. 9, pp. 2579–2605, Nov. 2008.

[49] Y. Zhai *et al.*, "MFFA-SARNET: Deep Transferred Multi-Level Feature Fusion Attention Network with Dual Optimized Loss for Small-Sample SAR ATR," *Remote Sens.*, vol. 12, no. 9, pp. 1385-1404, Apr. 2020.

[50] C. Zhang, S. Bengio, M. Hardt, B. Recht, and O. Vinyals, "Understanding Deep Learning Requires Rethinking Generalization," in *Proc. Int. Conf. Learn. Representations (ICLR)*, Toulon, France, Apr. 2017, pp. 1-15.

[51] T. Miyato, S. Maeda, M. Koyama, and S. Ishii, "Virtual Adversarial Training: A Regularization Method for Supervised and Semi-Supervised Learning," *IEEE Trans. Pattern Anal. Mach. Intell.*, vol. 41, no. 8, pp. 1979-1993, 2019.

[52] D.P. Kingma and M. Welling, "Auto-Encoding Variational Bayes," *in Proc. Int. Conf. Learn. Representations (ICLR)*, Banff, AB, Canada, Apr. 2014, pp. 412-422.